\documentclass[conference]{IEEEtran}
\IEEEoverridecommandlockouts
\usepackage{cite}
\usepackage{amsmath,amssymb,amsfonts}
\usepackage{algorithmic}
\usepackage{graphicx}
\usepackage{textcomp}
\usepackage{xcolor}
\usepackage{booktabs} 
\usepackage{hyperref}
\usepackage{makecell}
\usepackage{braket}

\renewcommand{\figureautorefname}{Figure~\negthinspace}

\renewcommand{\tableautorefname}{Table~\negthinspace}

\def\BibTeX{{\rm B\kern-.05em{\sc i\kern-.025em b}\kern-.08em
    T\kern-.1667em\lower.7ex\hbox{E}\kern-.125emX}}
\begin{document}

\title{Quantum Long Short-term Memory with Differentiable Architecture Search
\thanks{The views expressed in this article are those of the authors and do not represent the views of Wells Fargo. This article is for informational purposes only. Nothing contained in this article should be construed as investment advice. Wells Fargo makes no express or implied warranties and expressly disclaims all legal, tax, and accounting implications related to this article.}
}


\author{
\IEEEauthorblockN{
    Samuel Yen-Chi Chen\IEEEauthorrefmark{1}, 
    Prayag Tiwari\IEEEauthorrefmark{2}
}
\IEEEauthorblockA{\IEEEauthorrefmark{1} Wells Fargo, New York, NY, USA}
\IEEEauthorblockA{\IEEEauthorrefmark{2} School of Information Technology, Halmstad University, Sweden}
}

\maketitle

\begin{abstract}
Recent advances in quantum computing and machine learning have given rise to quantum machine learning (QML), with growing interest in learning from sequential data. Quantum recurrent models like QLSTM are promising for time-series prediction, NLP, and reinforcement learning. However, designing effective variational quantum circuits (VQCs) remains challenging and often task-specific. To address this, we propose DiffQAS-QLSTM, an end-to-end differentiable framework that optimizes both VQC parameters and architecture selection during training. Our results show that DiffQAS-QLSTM consistently outperforms handcrafted baselines, achieving lower loss across diverse test settings. This approach opens the door to scalable and adaptive quantum sequence learning.
\end{abstract}


\section{Introduction}
Quantum computing (QC) promises substantial speedups over classical computation for specific problem classes \cite{nielsen2010quantum}. In parallel, the development of artificial intelligence and machine learning (AI/ML) techniques has achieved remarkable success across diverse application domains. The emerging field of quantum machine learning (QML) seeks to harness the synergistic power of both QC and ML.
At the heart of many QML models lie variational quantum algorithms (VQAs) \cite{bharti2022noisy,cerezo2021variational}, which constitute the foundation of hybrid quantum-classical learning systems. These algorithms underpin the design of quantum neural networks (QNNs), which have been applied to a wide range of tasks including classification \cite{mitarai2018quantum,schuld2020circuit}, sequence learning \cite{chen2022quantumLSTM,li2023pqlm}, and reinforcement learning \cite{chen2020QRL,chen2023quantumLSTM_RL,meyer2022survey}.
Among these, sequence learning is a particularly critical application, with use cases spanning time-series forecasting, speech recognition, and natural language processing. Quantum long short-term memory (QLSTM) networks represent a prominent class of QML models tailored for such tasks. QLSTM has demonstrated competitive performance across several domains, including time-series prediction \cite{chen2022quantumLSTM} and quantum-enhanced language modeling \cite{li2023pqlm,di2022dawn,stein2023applying}.
Despite these advances, a key limitation remains: the quantum circuit architectures embedded within QLSTM modules are typically hand-crafted, requiring significant domain expertise and often tailored to specific problem characteristics. This design dependency poses a barrier to broader adoption, particularly for practitioners lacking a background in quantum information science.
To address this challenge, we propose a differentiable quantum architecture search framework (DiffQAS) integrated into the QLSTM model. This approach enables end-to-end training of both the conventional circuit parameters and the architectural control parameters that determine the contribution of candidate variational quantum circuits. Through comprehensive numerical experiments, we show that the resulting DiffQAS-QLSTM framework outperforms baseline QLSTM models with manually designed circuits on benchmark time-series prediction tasks.
We envision that this framework will facilitate the adoption of QML models, especially quantum sequence learners, by a broader range of domain experts, bridging the gap between quantum algorithm design and practical applications.
\section{Related Works}
\label{sec:related_works}
Modeling data with sequential or temporal dependencies, such as in time-series prediction and natural language processing (NLP), is a fundamental task in machine learning. To address such challenges within the quantum domain, quantum long short-term memory (QLSTM) networks have been proposed as a quantum analog of classical LSTM architectures \cite{chen2022quantumLSTM}. QLSTM models have shown empirical advantages over their classical counterparts under specific conditions. These models have been successfully applied to a wide range of applications, including time-series forecasting \cite{chen2022quantumLSTM}, NLP tasks \cite{li2023pqlm,di2022dawn,stein2023applying}, reinforcement learning in partially observable environments \cite{chen2023quantumLSTM_RL,chen2024quantumLSTM_RC_RL}, and quantum generative modeling \cite{chu2025lstm}. Despite these promising results, a key challenge remains: the construction of effective QLSTM architectures requires substantial quantum expertise, particularly in designing high-performance variational quantum circuits. This limits accessibility and scalability, especially for researchers and practitioners outside the quantum computing community.
\emph{Quantum architecture search} (QAS) \cite{martyniuk2024quantum} is an emerging research direction focused on automating the design of quantum circuit architectures tailored for specific quantum computational tasks such as QML. A variety of approaches have been proposed to tackle this challenge, with notable efforts rooted in evolutionary computation and reinforcement learning. Evolutionary QAS methods treat quantum circuit configurations as individuals in a population, where each architecture is encoded as a chromosome that undergoes mutation and selection. These approaches have been shown to be effective in discovering performant quantum neural networks (QNNs) for classification tasks \cite{altares2021automatic,altares2024autoqml}, reinforcement learning environments \cite{ding2022evolutionary}, and even for optimizing information-theoretic properties such as high effective dimension \cite{chen2024evolutionary}. In parallel, RL-based QAS methods have been applied to a wide range of tasks, including quantum state synthesis \cite{kuo2021quantum,ye2021quantum,sogabe2022model,zhu2023quantum}, solving machine learning problems \cite{dai2024quantum,wang2024rnn,rapp2024reinforcement}, estimating ground states in quantum chemistry \cite{kundu2025tensorrl,patel2024curriculum}, and addressing general optimization objectives \cite{fodera2024reinforcement}. While RL-based frameworks offer theoretical generality and policy-driven exploration, they often suffer in practice from challenges such as high sample complexity, optimizer sensitivity, and the inherent exploration–exploitation dilemma. A practical advantage of evolutionary approaches lies in their ease of parallelization, which makes them particularly suitable for deployment on high-performance computing (HPC) clusters. This allows for large-scale batch evaluations of candidate architectures. Nevertheless, both RL-based and evolutionary methods rely on discrete architecture search spaces, which quickly become intractable due to the combinatorial growth of candidate configurations as the search space expands. To overcome these limitations, recent studies have introduced differentiable QAS techniques \cite{zhang2022differentiable,sun2023differentiable,chen2024differentiable,chen2025learning_to_measure}, inspired by classical neural architecture search (NAS) frameworks such as DARTS \cite{liu2018darts}. These approaches relax the discrete architecture space into a continuous domain, typically by assigning trainable weights or distributions over a set of candidate substructures. This relaxation enables gradient-based optimization of both architecture and circuit parameters in a unified, end-to-end differentiable framework. In this work, we extend this paradigm by integrating differentiable QAS into the QLSTM model. This integration allows us to efficiently explore a significantly larger architecture space without incurring the sampling inefficiencies associated with RL or evolutionary methods. Moreover, our framework supports simultaneous optimization of variational circuit parameters and structural weights, facilitating smooth, gradient-driven training of quantum-enhanced models for sequence learning tasks.
\section{Quantum Neural Networks}
\begin{figure}[htbp]
\begin{center}
\includegraphics[width=1\columnwidth]{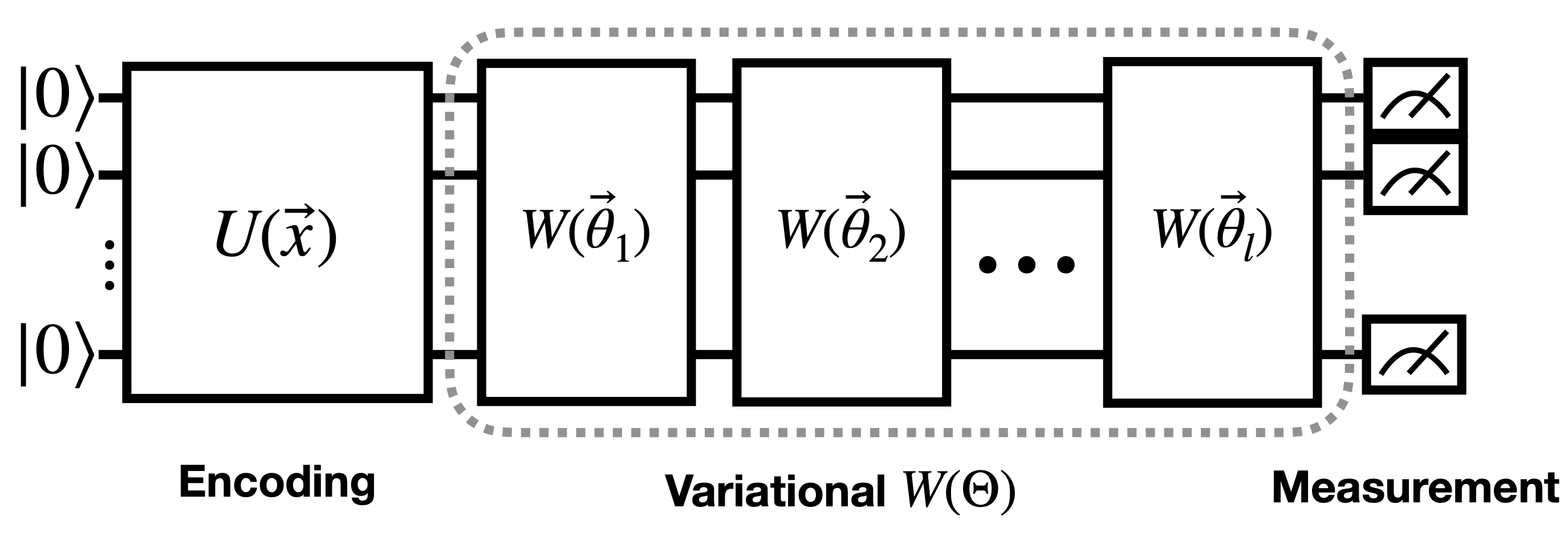}
\caption{{\bfseries Quantum Neural Network (QNN).}}
\label{fig:QNN}
\end{center}
\vskip -0.15in
\end{figure}
A typical quantum neural network (QNN), as shown in \figureautorefname{\ref{fig:QNN}}, consists of three key stages: a data encoding circuit $U(\vec{x})$ that embeds classical inputs into quantum states, a trainable variational circuit $W(\Theta)$, and a final measurement using observables $\hat{B}_k$. The input-dependent circuit prepares the initial state $U(\vec{x})\ket{0}^{\otimes n}$, where $\vec{x}$ is the classical feature vector and $n$ is the number of qubits.
This encoded state is then transformed by the variational circuit, yielding the final quantum state, $\ket{\Psi} = W(\Theta) U(\vec{x})\ket{0}^{\otimes n}$,
where the variational circuit is composed of $M$ layers of trainable unitary blocks, $W(\Theta)=  \prod_{j = M}^{1} W_{j}(\vec{\theta_{j}}), \quad \Theta = \{\vec{\theta_{1}} \cdots \vec{\theta_{M}}\}$.
Information is retrieved from the VQC by measuring Hermitian observables $\hat{B}_k$, producing expectation values $\langle \hat{B}_k \rangle = \bra{\Psi} \hat{B}_k \ket{\Psi}$. Practically, these values are estimated either through repeated measurements (i.e., sampling over multiple shots) on real quantum computers or computed analytically in simulation software. The VQC can thus be viewed as a quantum function $\vec{f}(\vec{x}; \Theta) = (\langle \hat{B}_1 \rangle, \dots, \langle \hat{B}_K \rangle)$, where observables $\hat{B}_k$ determine how quantum information is projected into the classical domain.
\section{Quantum LSTM}
\begin{figure}[htbp]
\begin{center}
\includegraphics[width=1\columnwidth]{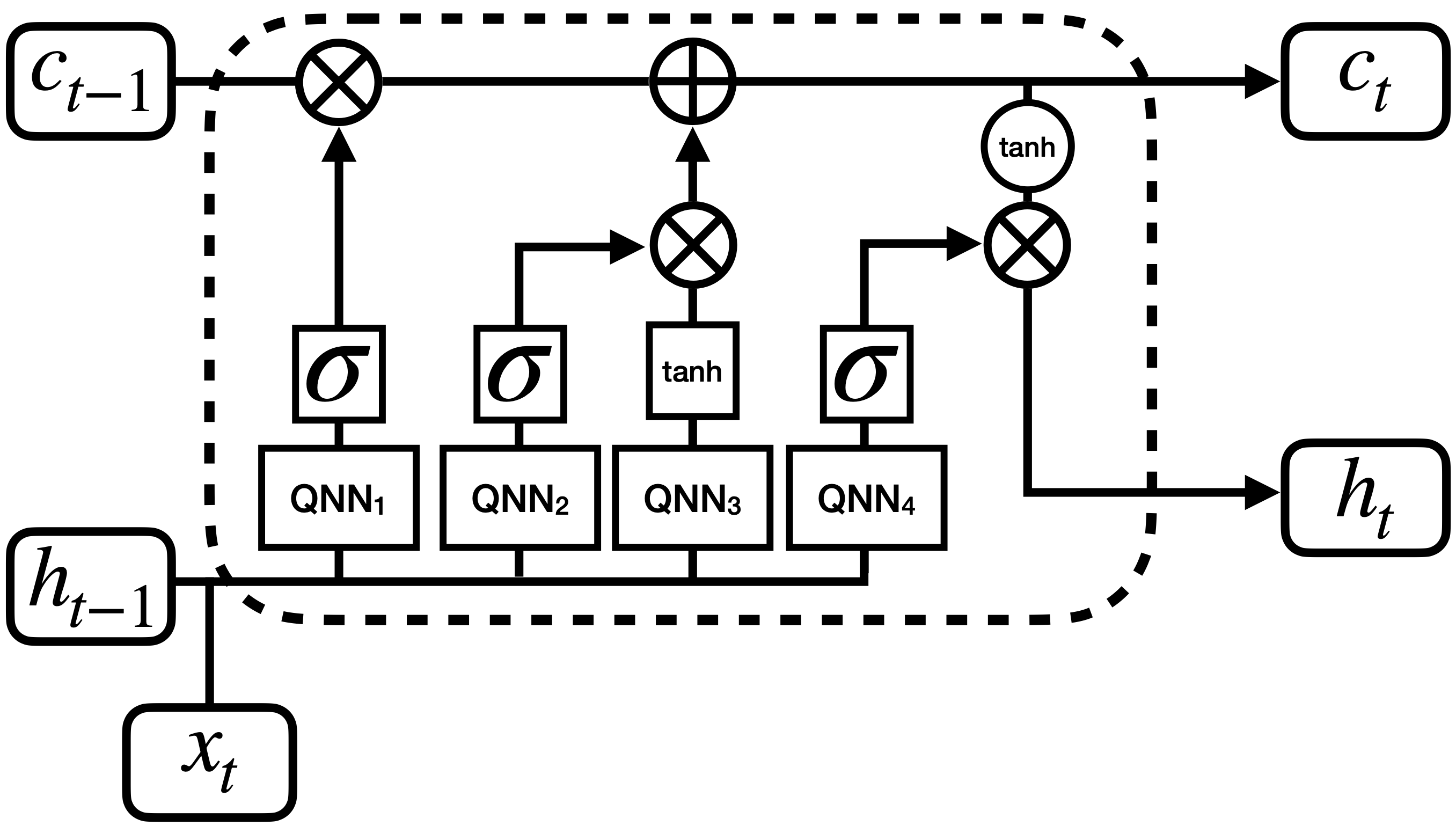}
\caption{{\bfseries Quantum Long Short-term Memory (QLSTM).}}
\label{fig:QLSTM}
\end{center}
\end{figure}

The Quantum Long Short-Term Memory (QLSTM) model extends the classical LSTM architecture by replacing classical neural network components with QNNs \cite{chen2022quantumLSTM}, as illustrated in \figureautorefname{\ref{fig:QLSTM}}.
A formal definition of a QLSTM cell is given by,
\begin{subequations}
\allowdisplaybreaks
    \begin{align}
    f_{t} &= \sigma\left(\text{QNN}_{1}(v_t)\right) \label{eqn:qlstm-f}\\
    i_{t} &= \sigma\left(\text{QNN}_{2}(v_t)\right) \label{eqn:qlstm-i}\\ 
    \tilde{C}_{t} &= \tanh \left(\text{QNN}_{3}(v_t)\right) \label{eqn:qlstm-bigC}\\
    c_{t} &= f_{t} * c_{t-1} + i_{t} * \tilde{C}_{t} \label{eqn:qlstm-c}\\
    o_{t} &= \sigma\left(\text{QNN}_{4}(v_t)\right) \label{eqn:qlstm-o}\\ 
    h_{t} &= o_{t} * \tanh \left(c_{t}\right)\label{eqn:qlstm-h}
    \end{align}
    \label{eqn:qlstm}
\end{subequations}
where $v_t=\left[h_{t-1} x_{t}\right]$ represents the concatenation of input $x_{t}$ at time-step $t$ and the hidden state $h_{t-1}$ from the previous time-step $t-1$. 
Previous studies have shown that QLSTM can outperform classical LSTM models on various time-series prediction tasks, particularly when both models have comparable numbers of trainable parameters \cite{chen2022quantumLSTM}. Beyond time-series forecasting, QLSTM has also demonstrated promising performance in NLP \cite{li2023pqlm,di2022dawn,stein2023applying} and in RL under partial observability \cite{chen2023quantumLSTM_RL}. However, current QLSTM implementations typically depend on quantum experts or trial-and-error procedures to manually design the underlying QNN architectures, limiting accessibility and scalability.
\section{Differentiable Quantum Architecture Search}
\begin{figure}[htbp]
\begin{center}
\includegraphics[width=1\columnwidth]{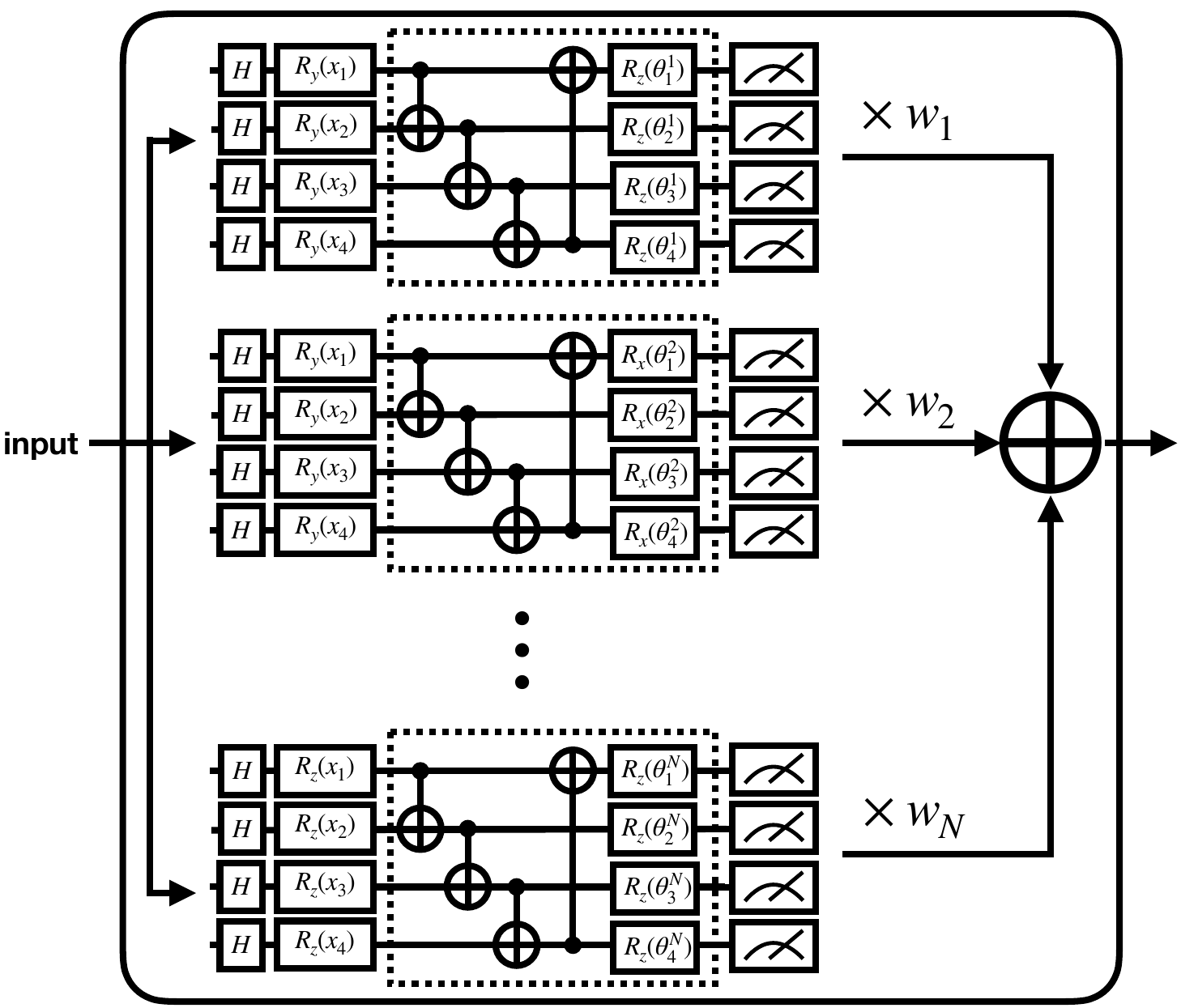}
\caption{{\bfseries Differentiable Quantum Architecture Search (DiffQAS) Block.}}
\label{fig:DiffQAS_Block}
\end{center}
\end{figure}

\begin{figure}[htbp]
\vskip -0.1in
\begin{center}
\includegraphics[width=1\columnwidth]{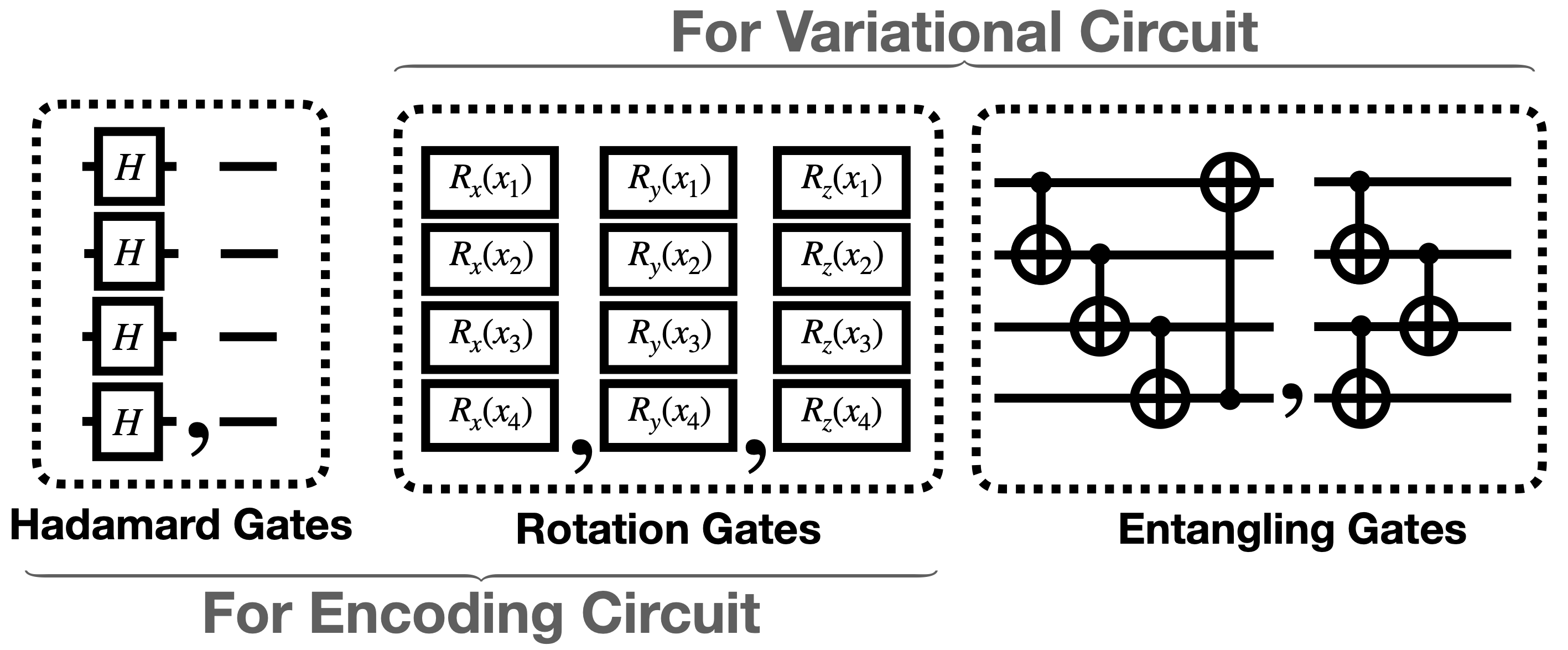}\vskip -0.1in
\caption{{\bfseries Ansatzes of QNN/VQC considered in this work.}}
\label{fig:Candidate_Circuits}
\end{center}
\vskip -0.2in
\end{figure}

Inspired by classical neural architecture search (NAS) techniques~\cite{liu2018darts} and recent advances in differentiable quantum architecture search (DiffQAS)~\cite{zhang2022differentiable,chen2024differentiable}, our approach begins by defining a set of candidate quantum subcircuits that serve as building blocks for the architecture search space.
Consider a quantum circuit $\mathcal{C}$ composed of a sequence of modular units $\mathcal{S}_{1}, \mathcal{S}_{2}, \dots, \mathcal{S}_{n}$. Each module $\mathcal{S}_{i}$ is selected from a predefined set of candidate subcircuits $\mathcal{B}_{i}$, where the size of $\mathcal{B}_{i}$ determines the number of available design choices at that location. The overall architecture search space for $\mathcal{C}$ thus comprises $N = |\mathcal{B}_{1}| \times |\mathcal{B}_{2}| \times \cdots \times |\mathcal{B}_{n}|$ distinct circuit configurations.
Each candidate configuration $\mathcal{C}_{j}$, where $j \in \{1, \dots, N\}$, is assigned a learnable structural weight  $w_j$. In addition, every configuration $\mathcal{C}_{j}$ can be  equipped with its own set of trainable quantum parameters $ \theta_j$. Depending on the task at hand, certain configurations may yield strong performance after training, while others may fail to generalize effectively.
As illustrated in \figureautorefname{\ref{fig:DiffQAS_Block}}, we define an ensemble function $f_{\mathcal{C}}$ as a weighted combination over all $N$ candidates to incorporate effects from all circuit realizations, $f_{\mathcal{C}} = \sum_{j = 1}^{N} w_{j}f_{\mathcal{C}_{j}}$, where for brevity, we omit explicit references to the data inputs $\vec{x}$ and circuit parameters $\theta_j$.
The ensemble output $f_{\mathcal{C}}$ is then used by a task-specific loss function $\mathcal{L}(f_{\mathcal{C}})$. Through automatic differentiation, gradients with respect to the structural weights $w_j$, i.e., $\nabla_{w_j} \mathcal{L}(f_{\mathcal{C}})$, can be computed, enabling efficient updates via standard gradient-based optimization methods.
Within the DiffQAS framework, variational quantum circuits (VQCs) are constructed by selecting from a set of pre-defined ansatz candidates, as depicted in \figureautorefname{\ref{fig:Candidate_Circuits}}. 
For instance, consider the design space of an \emph{encoding circuit} that offers two options for initialization (e.g., with or without a Hadamard gate) and three choices of rotation gates for encoding input features, resulting in $2 \times 3 = 6$ possible configurations. Similarly, the \emph{variational circuit} may be constructed by selecting from two entanglement patterns and three types of rotation gates for parameterized layers, again yielding $2 \times 3 = 6$ variants. The overall architecture space in this scenario thus comprises $6 \times 6 = 36$ unique circuit realizations.
In the proposed DiffQAS-QLSTM framework, we incorporate the DiffQAS mechanism into the original QLSTM architecture by replacing the manually designed VQC or QNN module with a flexible \emph{DiffQAS Block}, as illustrated in \figureautorefname{\ref{fig:DiffQAS_Block}}.
\begin{figure}[htbp]
\begin{center}
\includegraphics[width=1\columnwidth]{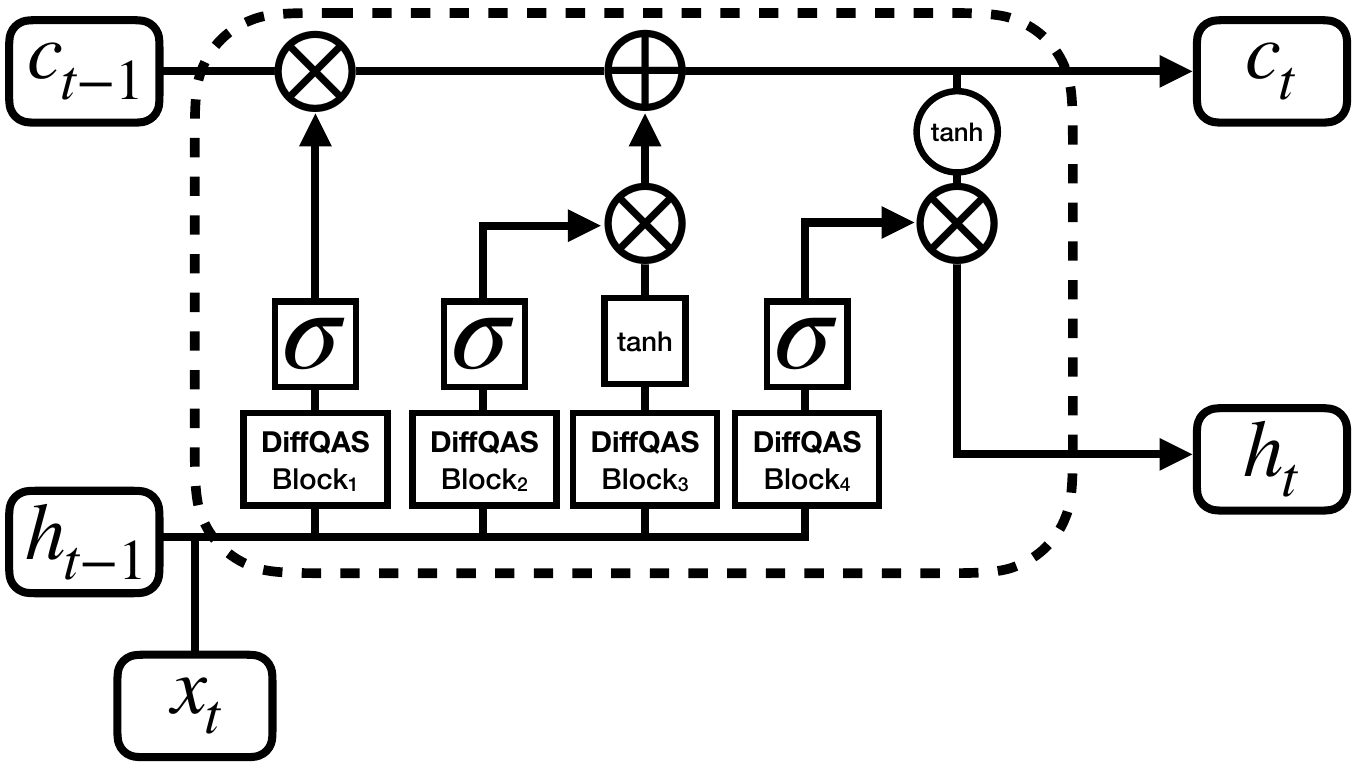}
\caption{{\bfseries QLSTM with DiffQAS.}}
\label{fig:DiffQAS_QLSTM}
\end{center}
\vskip -0.15in
\end{figure}
\section{Numerical Results and Discussions}
\begin{figure}[htbp]
\begin{center}
\includegraphics[width=1\columnwidth]{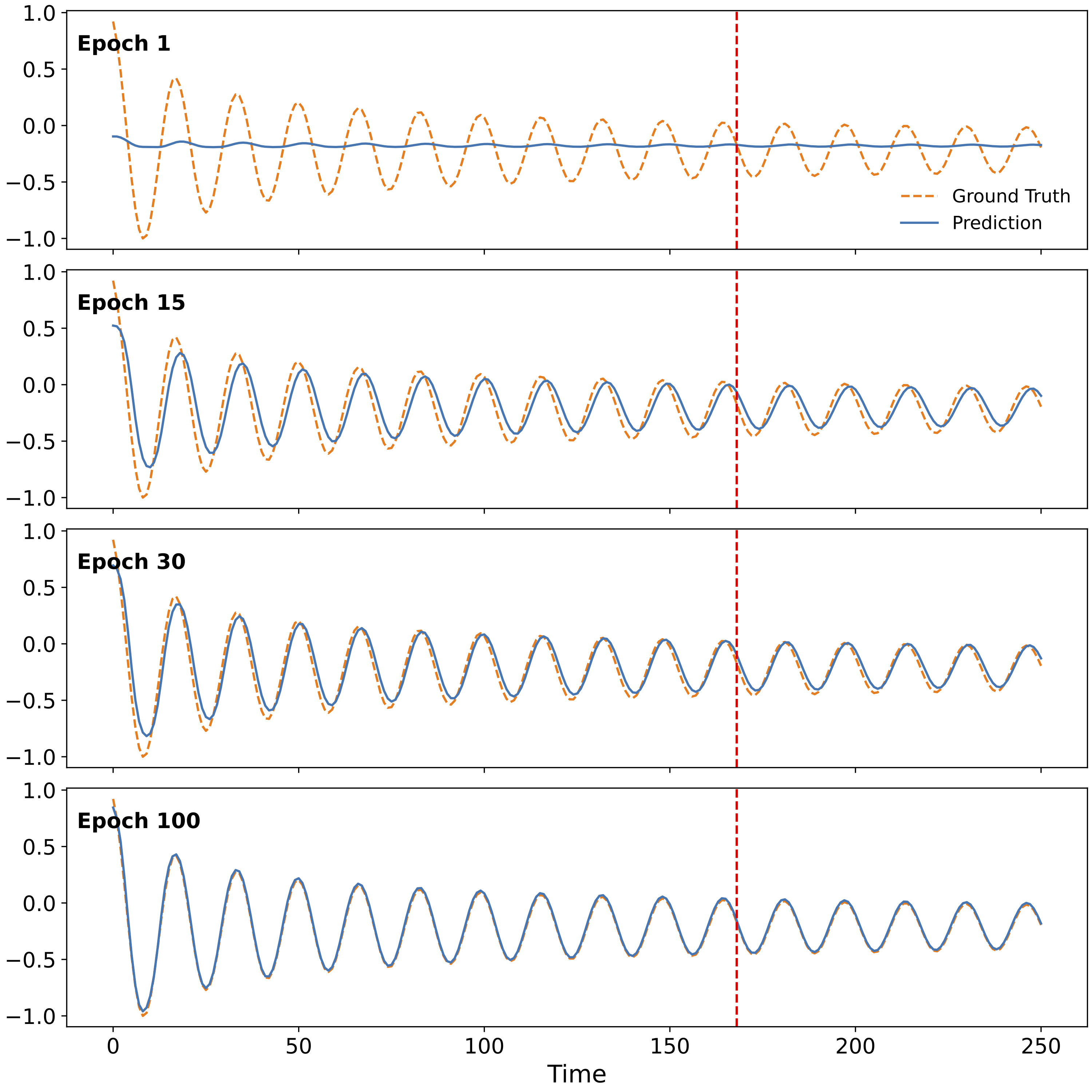}
\caption{{\bfseries Time-series predictions of DiffQAS-QLSTM (NonShared) on the Bessel function $J_2$ task at different training epochs.} Each panel shows the model’s prediction (solid blue) compared to the ground truth (dashed orange). The vertical red dashed line indicates the transition between the training portion (first two-thirds of the dataset) and the testing portion (last one-third). As training progresses, the model increasingly captures the underlying dynamics and achieves stable predictions throughout the testing segment.}
\label{fig:results_DiffQAS_NonShared_Bessel}
\end{center}
\vskip -0.15in
\end{figure}

\begin{figure}[htbp]
\begin{center}
\includegraphics[width=1\columnwidth]{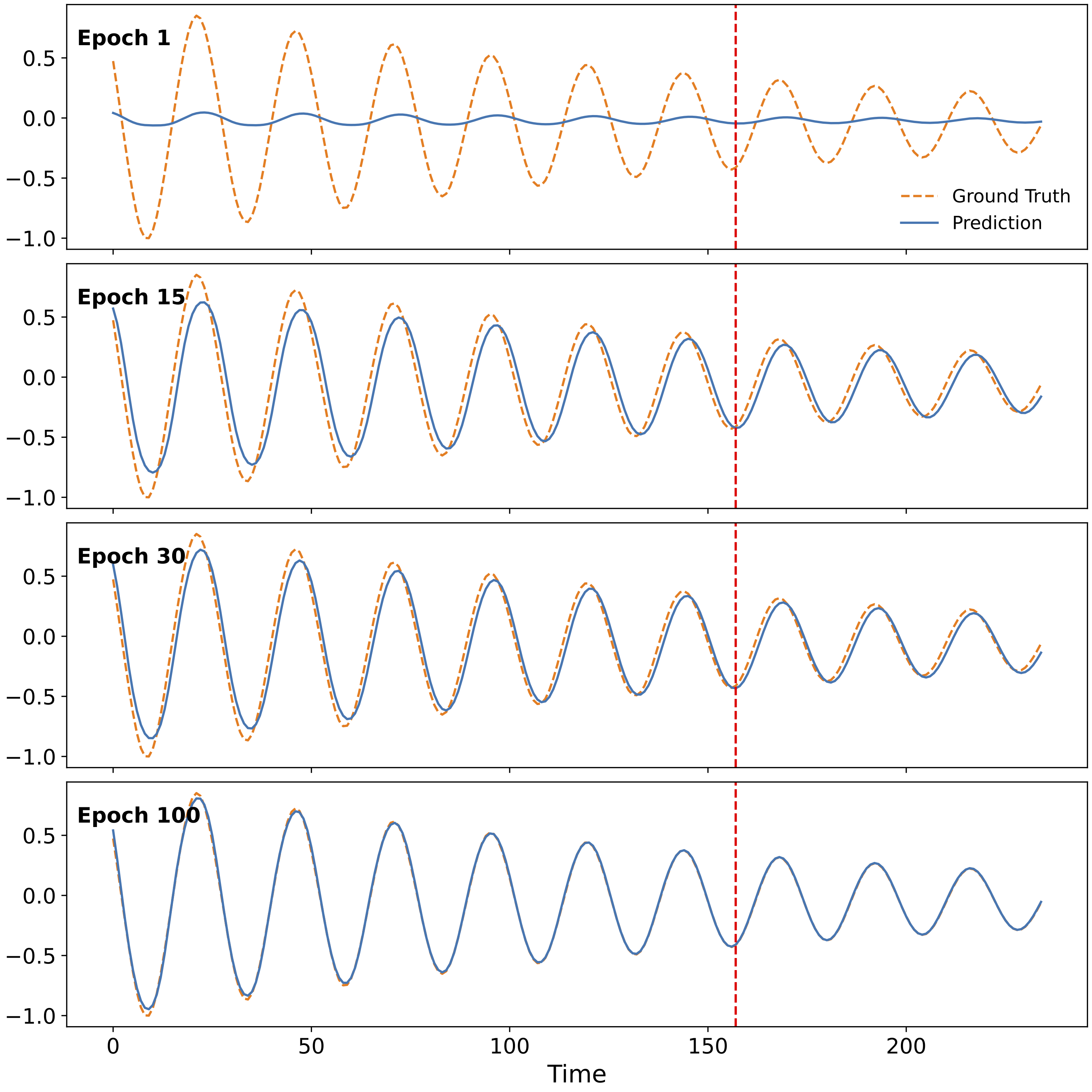}
\caption{{\bfseries Time-series predictions of DiffQAS-QLSTM (NonShared) on the Damped SHM task at different training epochs.} The model quickly captures the damped oscillatory pattern and maintains accurate predictions in testing region.}
\label{fig:results_DiffQAS_NonShared_DampedSHM}
\end{center}
\vskip -0.15in
\end{figure}

\begin{figure}[htbp]
\begin{center}
\includegraphics[width=1\columnwidth]{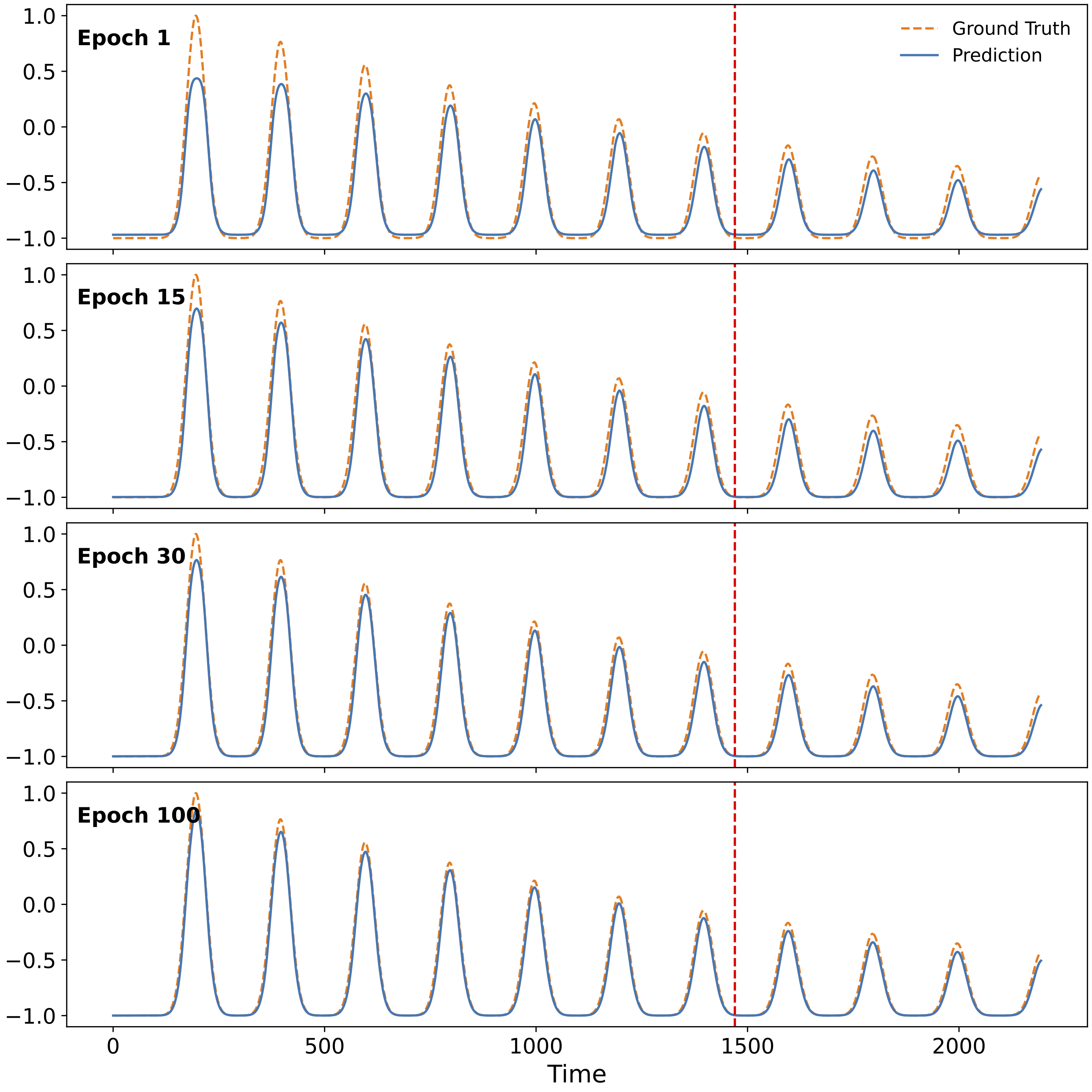}
\caption{{\bfseries Time-series predictions of DiffQAS-QLSTM (NonShared) on the Delayed Quantum Control task at different training epochs.} Despite the task’s delayed feedback dynamics, the model remains stable throughout the prediction horizon.}
\label{fig:results_DiffQAS_NonShared_Delayed_Quantum_Control}
\end{center}
\vskip -0.15in
\end{figure}

\begin{figure}[htbp]
\begin{center}
\includegraphics[width=1\columnwidth]{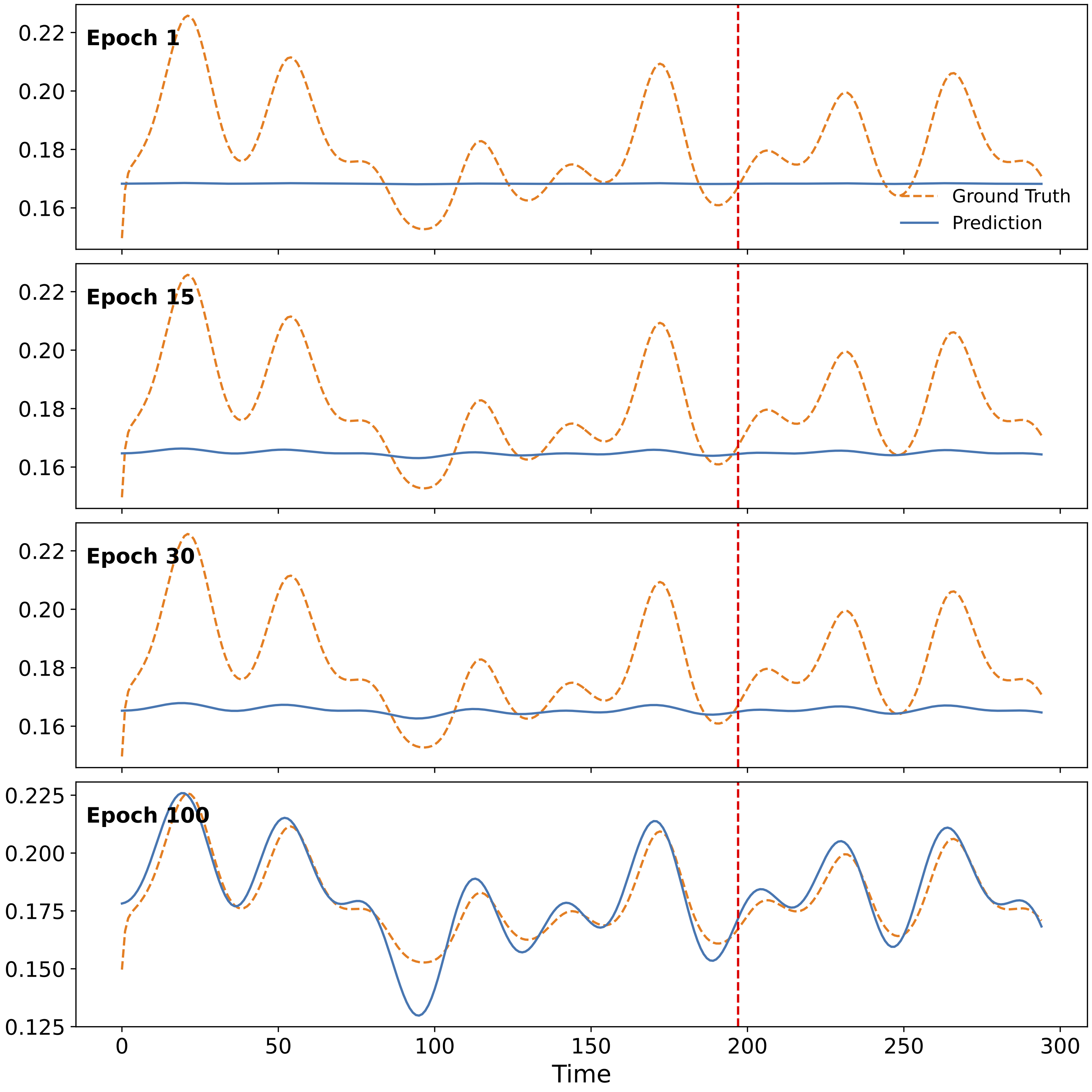}
\caption{{\bfseries Time-series predictions of DiffQAS-QLSTM (NonShared) on the NARMA 5 task at different training epochs.} Same visualization setup as in \figureautorefname{\ref{fig:results_DiffQAS_NonShared_Bessel}}. The model gradually learns the nonlinear temporal dependencies and achieves stable predictions as training progresses.}
\label{fig:results_DiffQAS_NonShared_NARMA_5}
\end{center}
\vskip -0.15in
\end{figure}

\begin{figure}[htbp]
\begin{center}
\includegraphics[width=1\columnwidth]{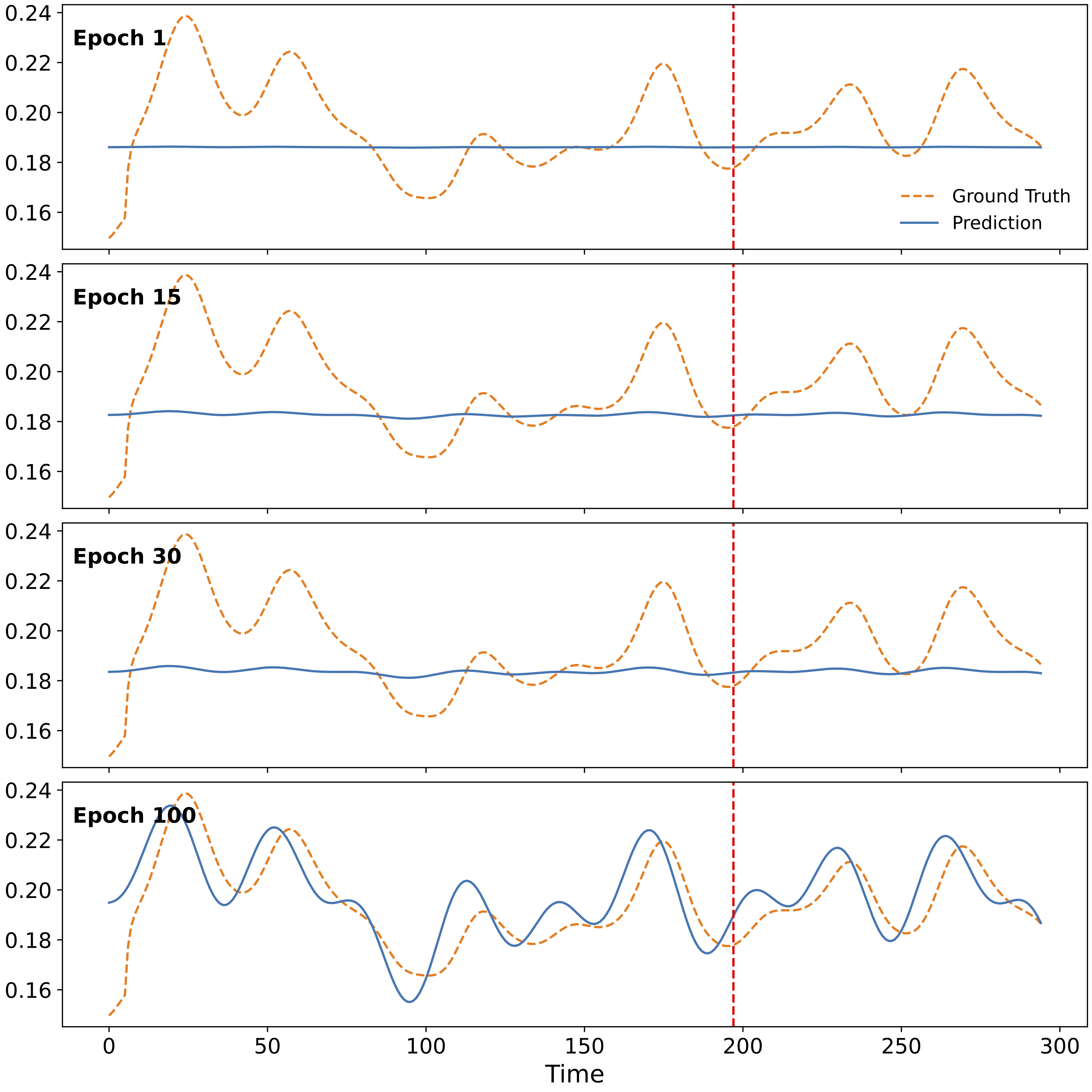}
\caption{{\bfseries Time-series predictions of DiffQAS-QLSTM (NonShared) on the NARMA 10 task at different training epochs.} The model maintains prediction stability even in the presence of high-order temporal dependencies.}
\label{fig:results_DiffQAS_NonShared_NARMA_10}
\end{center}
\vskip -0.15in
\end{figure}

We evaluate the performance of the proposed DiffQAS-QLSTM model under the following four configurations:
(1) \emph{NonShared} — each candidate circuit maintains its own trainable parameter set;
(2) \emph{Shared} — all candidate circuits share a single trainable parameter set;
(3) \emph{Reservoir-NonShared} — each candidate circuit is assigned a distinct, randomly initialized parameter set that remains fixed during training;
(4) \emph{Reservoir-Shared} — all candidates share the same randomly initialized, non-trainable parameter set.
For comparison, we include baseline models based on six manually designed QNN architectures, following the DiffQAS study in \cite{chen2024differentiable}, as summarized in \tableautorefname{\ref{tab:different_circuit_baselines}}.
The training and evaluation protocol follows the methodology described in~\cite{chen2022quantumLSTM,chen2022reservoir,chen2024QFWP}. Specifically, the model is trained to predict the $(N+1)$-th value in a sequence given the preceding $N$ observations. For example, at time step $t$, the input to the model is $[x_{t-4}, x_{t-3}, x_{t-2}, x_{t-1}]$ (with $N=4$), and the target output is $y_t$, which should approximate the ground truth $x_t$. Throughout all time-series experiments, we fix the sequence length parameter to $N=4$.
To further evaluate the learning behavior and  forecasting capability of the proposed DiffQAS-QLSTM (NonShared), we visualize its predictions across training epochs on five representative time-series benchmarks: Bessel (\figureautorefname{\ref{fig:results_DiffQAS_NonShared_Bessel}}), Damped SHM (\figureautorefname{\ref{fig:results_DiffQAS_NonShared_DampedSHM}}), Delayed Quantum Control (\figureautorefname{\ref{fig:results_DiffQAS_NonShared_Delayed_Quantum_Control}}), NARMA 5 (\figureautorefname{\ref{fig:results_DiffQAS_NonShared_NARMA_5}}), and NARMA 10 (\figureautorefname{\ref{fig:results_DiffQAS_NonShared_NARMA_10}}). The vertical red dashed line indicating the boundary between the training part and the testing part.
\begin{table}[htbp]
\caption{VQC baselines.}
\label{tab:different_circuit_baselines}
\begin{tabular}{|l|l|l|l|l|l|l|}
\hline
\diaghead{\theadfont Diag ColumnmnHead II}%
  {Component}{VQC config}                        & 1 & 2 & 3 & 4 & 5 & 6 \\ \hline
Encoding                & $R_{y}$  & $R_{z}$  & $R_{z}$  & $R_{y}$  & $R_{x}$  & $R_{x}$  \\ \hline
Trainable Rotation Gate & $R_{y}$  & $R_{y}$  & $R_{z}$  & $R_{z}$  & $R_{z}$  & $R_{y}$  \\ \hline
\end{tabular}
\end{table}
Across all tasks, the model progressively improves its predictive accuracy and stability as training advances. In the Bessel and Damped SHM tasks, which involve smooth oscillatory dynamics, the model quickly learns the underlying structure and achieves near-perfect prediction by epoch 30. On the more challenging Delayed Quantum Control and NARMA tasks, the model still demonstrates strong long-term stability and accuracy, despite delayed feedback and high-order dependencies. Notably, in the NARMA 10 task, DiffQAS-QLSTM exhibits robustness in capturing intricate temporal correlations, maintaining consistent alignment with ground truth over extended horizons.

These results highlight the effectiveness of DiffQAS in guiding the architecture design of quantum-enhanced recurrent models, enabling fast convergence and robust generalization across a variety of temporal dynamics.
In contrast, manually designed QLSTM baselines (Config 1–6) often exhibit suboptimal performance, with higher prediction drift or instability-particularly on more complex tasks such as Damped SHM, Delayed Quantum Control and NARMA 10. As shown in \tableautorefname{\ref{table:all_results_compare}}, DiffQAS-QLSTM (NonShared) consistently achieves the lowest test MSE on most datasets, outperforming the best baseline by a clear margin.
We further conduct ablation studies to isolate the contribution of structural discovery and parameter optimization in DiffQAS-QLSTM. In the Reservoir setting, the variational quantum circuit (VQC) architecture is discovered via DiffQAS but its parameters are randomly initialized and fixed throughout training. Only the structural components-such as the structural weights $w_{j}$ or output projection layers-are updated. Despite benefiting from optimized structures, the reservoir variants exhibit significantly degraded performance on most tasks (see \tableautorefname{\ref{table:all_results_compare}}), highlighting that structural expressivity alone is insufficient without adaptive parameter learning.

In contrast, the Shared variant allows different tasks to use the same parameters across sampled VQCs, while the NonShared variant assigns unique trainable parameters to each structure. As shown in both the rollout predictions and final test MSE, the NonShared configuration consistently outperforms all others, underscoring the importance of both flexible architecture selection and fully trainable quantum parameters.
\section{Conclusion}
We proposed DiffQAS-QLSTM, a quantum sequence learning framework that integrates differentiable quantum architecture search into QLSTM models. By enabling end-to-end optimization over both circuit parameters and architecture configurations, our method outperforms manually designed baselines across time-series prediction tasks considered. This work paves the way for more accessible and adaptable quantum neural architectures in sequential learning applications.
\begin{table*}[ht]
\centering
\caption{Final Test MSE of QLSTM Variants Across Time-Series Datasets (Lower is Better)}
\begin{tabular}{lccccc}
\toprule
\textbf{Model} & \textbf{Bessel} $\downarrow$ & \textbf{Damped SHM} $\downarrow$ & \textbf{Delayed Quantum Control} $\downarrow$ & \textbf{NARMA 5} $\downarrow$ & \textbf{NARMA 10} $\downarrow$ \\
\midrule
\textbf{DiffQAS-QLSTM-NonShared} & 0.000229 & \textbf{0.000019} & \textbf{0.001859} & 0.000030 & 0.000094 \\
DiffQAS-QLSTM-Shared & \textbf{0.000117} & 0.000036 & 0.002486 & 0.000472 & 0.000385 \\
DiffQAS-QLSTM-Reservoir-NonShared & 0.006529 & 0.007489 & 0.003567 & \textbf{0.000016} & \textbf{0.000074} \\
DiffQAS-QLSTM-Reservoir-Shared & 0.006546 & 0.010023 & 0.004227 & 0.000494 & 0.000412 \\
Config 1 & 0.001324 & 0.000601 & 0.005507 & 0.000032 & 0.000123 \\
Config 2 & 0.002768 & 0.003803 & 0.004591 & 0.000273 & 0.000188 \\
Config 3 & 0.023947 & 0.046938 & 0.084120 & 0.000273 & 0.000188 \\
Config 4 & 0.007316 & 0.010588 & 0.001931 & 0.000025 & 0.000101 \\
Config 5 & 0.023947 & 0.046938 & 0.084120 & 0.000273 & 0.000188 \\
Config 6 & 0.024024 & 0.046886 & 0.077084 & 0.000273 & 0.000188 \\
\bottomrule
\end{tabular}
\label{table:all_results_compare}
\end{table*}
%
\bibliographystyle{IEEEtran}
\bibliography{bib/qas,bib/qml_examples,bib/qc}

\end{document}